%% file: acl_latex.tex
\pdfoutput=1

\documentclass[11pt]{article}
\usepackage[preprint]{acl}

\usepackage{times}

\usepackage[T1]{fontenc}

\usepackage[utf8]{inputenc}

\usepackage{microtype}
\usepackage{subcaption}
\usepackage{inconsolata}
\usepackage{amsmath}
\usepackage{bbm}
\usepackage{dsfont}
\usepackage{graphicx}
\usepackage{booktabs}
\usepackage{array}
\usepackage{listings}
\usepackage{xcolor}  

\usepackage{multirow}
\usepackage{arydshln} 
\usepackage{graphicx}
\usepackage{subcaption}
\usepackage{float}
\usepackage{tabularx}
\usepackage{booktabs}
\usepackage{multirow}
\usepackage{pifont}
\usepackage{tcolorbox}
\usepackage{hyperref}
\usepackage{CJKutf8}

\definecolor{highlightcolor}{RGB}{255, 255, 204} 
\definecolor{myblue}{RGB}{59, 176, 240}

\title{LaTeXTrans: Structured LaTeX Translation \\ with Multi-Agent Coordination}

\author{
    Ziming Zhu\textsuperscript{\rm 1}\thanks{Authors contributed equally.}, 
    Chenglong Wang\textsuperscript{\rm 1}$^{*}$, 
    Haosong Xu\textsuperscript{\rm 1},
    Shunjie Xing\textsuperscript{\rm 1}, 
    Yifu Huo\textsuperscript{\rm 1}, \\
    \textbf{Fengning Tian\textsuperscript{\rm 2}, 
    Quan Du\textsuperscript{\rm 2},
    Di Yang\textsuperscript{\rm 1,2}, 
    Chunliang Zhang\textsuperscript{\rm 1,2},
    Tong Xiao\textsuperscript{\rm 1,2}\thanks{Corresponding author.}
    and
    Jingbo Zhu\textsuperscript{\rm 1,2}} \\
    \textsuperscript{\rm 1}School of Computer Science and Engineering, Northeastern University, Shenyang, China \\
    \textsuperscript{\rm 2}NiuTrans Research, Shenyang, China \\
    {\{zhuzm0721, clwang1119\}@gmail.com},
    {\{xiaotong, zhujingbo\}@mail.neu.edu.cn}
}

\begin{document}

\maketitle
\begin{abstract}
Despite the remarkable progress of modern machine translation (MT) systems on general-domain texts, translating structured LaTeX-formatted documents remains a significant challenge. These documents typically interleave natural language with domain-specific syntax, such as mathematical equations, tables, figures, and cross-references, all of which must be accurately preserved to maintain semantic integrity and compilability. 
In this paper, we introduce LaTeXTrans, a collaborative multi-agent system designed to address this challenge. LaTeXTrans ensures format preservation, structural fidelity, and terminology consistency through six specialized agents: 1) a \textit{Parser} that decomposes LaTeX into translation-friendly units via placeholder substitution and syntax filtering; 2) a \textit{Translator}, \textit{Validator}, \textit{Summarizer}, and \textit{Terminology Extractor} that work collaboratively to ensure context-aware, self-correcting, and terminology-consistent translations; 3) a \textit{Generator} that reconstructs the translated content into well-structured LaTeX documents.
Experimental results show that LaTeXTrans outperforms mainstream MT systems in both translation accuracy and structural preservation. The source code\footnote{\url{https://github.com/NiuTrans/LaTeXTrans}}, the online demonstration platform\footnote{\url{https://latextrans.online}}, and a demo video\footnote{\url{https://youtu.be/-ODRUTE-VU8}} are publicly available.
\end{abstract}


\section{Introduction}
LaTeX~\cite{urban1986introduction} is a widely used macro system built on TeX for typesetting complex and structured documents, and has become the standard infrastructure for scholarly publishing worldwide. Although LaTeX itself is language-agnostic, academic communication remains overwhelmingly dominated by English: nearly 98\% of scientific papers are published in English~\cite{kleidermacher}, while only about 5\% of the global population is proficient in the language as of 2025~\cite{ManySpeakEnglish}. This extreme asymmetry imposes a significant burden on most researchers. Non-native English speakers must not only master domain knowledge but also navigate linguistically demanding and structurally complex LaTeX documents in a foreign language.

To alleviate this burden, the most straightforward approach is to first compile LaTeX source files into a PDF and then apply machine translation directly at the PDF level (\textit{a.k.a.}  PDF translation).
However, this approach often results in incomplete formatting due to errors in PDF parsing. An alternative line of research advocates translating directly at the LaTeX source level before recompilation into the target language. Although prior work has demonstrated improved format preservation compared to PDF translation \cite{hoy2025latexmt}, such systems are typically limited to short and well-structured LaTeX source inputs. In practice, however, real-world LaTeX projects are rarely monolithic or syntactically simple. They often consist of multiple interdependent source files, user-defined commands, and domain-specific macros.


Given these limitations, there remains a strong demand for a translation system capable of handling real-world LaTeX projects in their full structural complexity. To this end, we propose LaTeXTrans, a collaborative multi-agent framework that directly translates raw LaTeX source files while preserving structural fidelity and semantic integrity. Unlike prior approaches that focus on isolated snippets, LaTeXTrans is designed to robustly handle multi-file architectures, user-defined commands, and macro-intensive environments. It consists of three modules and six specialized agents:

\begin{itemize}
    \vspace{-1.8mm}
    \item \textit{Parsing Module}: Responsible for fine-grained analysis of LaTeX-formatted documents. To handle the structural complexity of LaTeX projects, we design a \textit{Parser} agent equipped with a placeholder mechanism and a syntax filter, which together decompose the source into manageable translation units.
    \vspace{-2mm}
    \item \textit{Translation Module}: This module leverages a team of collaborative agents, including a \textit{Translator}, \textit{Validator}, \textit{Summarizer}, and \textit{Terminology Extractor}, which work together to perform context-aware and self-correcting translation of the parsed units.
    \vspace{-2mm}
    \item \textit{Generation Module}: A \textit{Generator} agent reconstructs the translated document by reinserting the translated content into the original LaTeX structure, producing well-formatted LaTeX source in the target language.
\end{itemize}

\begin{figure*}[!t]
  \includegraphics[width=\textwidth]{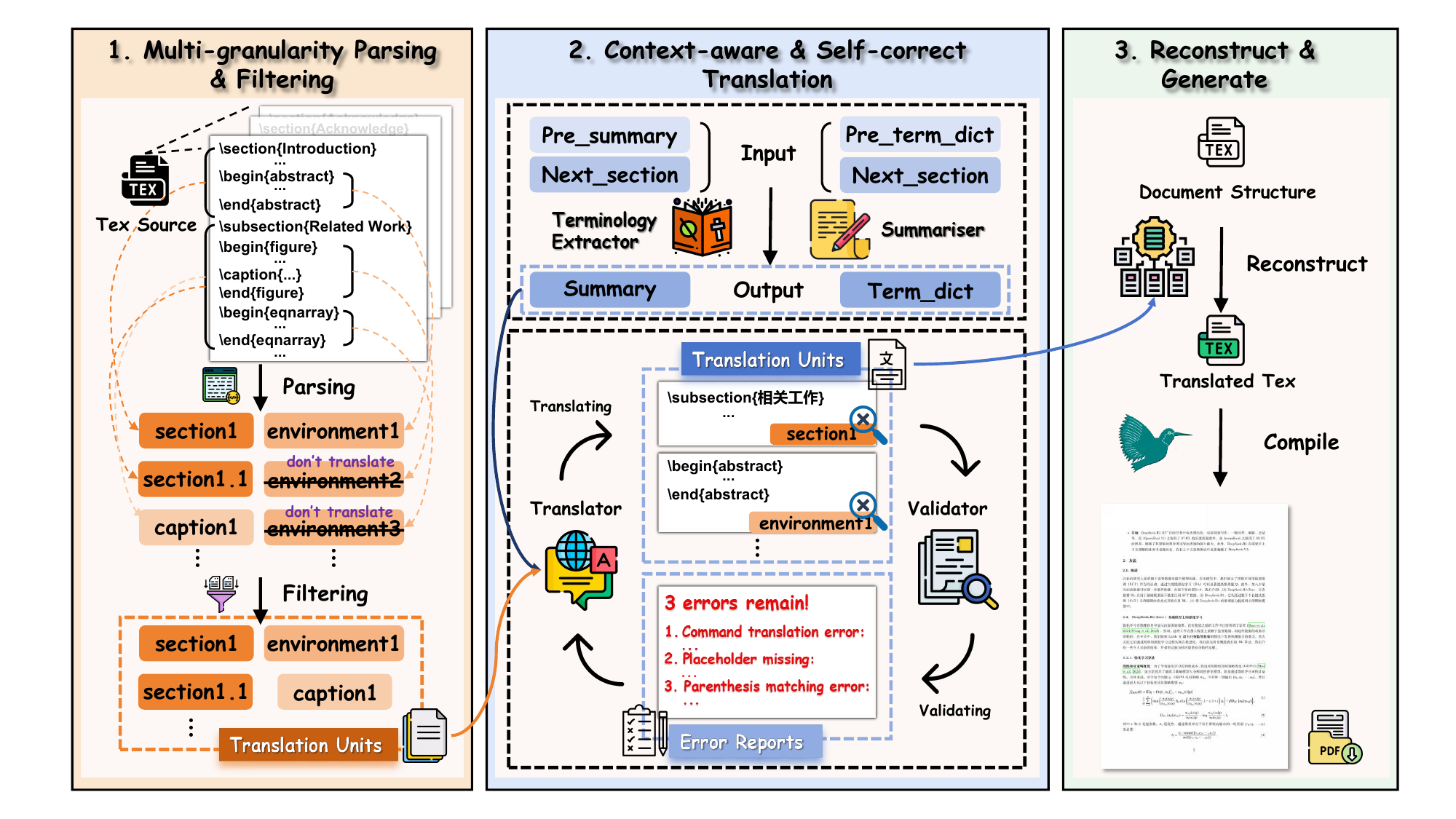}
  \caption{The architecture of our LaTeXTrans system.}
  \vspace{-0.3cm}
  \label{fig:LaTeXTrans}
\end{figure*}

To evaluate LaTeXTrans, we construct a multi-domain, multilingual benchmark for LaTeX source-level translation using real-world TeX projects from arXiv. We compare our system with representative baselines for formatted-text translation, assessing both structural preservation and translation accuracy. The results demonstrate that LaTeXTrans robustly handles complex LaTeX project architectures, maintains syntactic and structural consistency during translation, and consistently generates successfully compiled target-language PDFs. In the En–Zh setting, LaTeXTrans achieves a one-pass compilation success rate of 97\% across all test samples, with an average formatting error rate below 0.5 per document.


\section{Related works}

\paragraph{LLM-based Machine Translation.}
The emergence of large language models (LLMs) has introduced a new paradigm for machine translation, moving beyond traditional supervised training on parallel corpora toward more general-purpose language understanding \cite{gain2025bridginglinguisticdividesurvey}. Models such as GPT-3 \cite{brown2020languagemodelsfewshotlearners}, PaLM \cite{chowdhery2022palmscalinglanguagemodeling}, and GPT-4 exhibit strong multilingual capabilities without task-specific translation training. By leveraging in-context learning, LLM-based translation performs competitively in zero-shot and few-shot settings~\citep{vilar2023promptingpalmtranslationassessing,luo2025decoderonlylargelanguagemodels}, particularly for high-resource language pairs. Unlike traditional neural machine translation (NMT), which often requires domain- or language-specific retraining, LLMs demonstrate broader cross-task and cross-lingual generalization with minimal additional data.

\paragraph{Multi-Agent Systems.}
More recently, the emergence of LLMs has opened new possibilities for multi-agent systems (MAS). In LLM-based MAS, each agent is instantiated as an LLM-powered entity capable of natural language reasoning, planning, and collaboration. Systems such as AutoGPT~\citep{yang2023autogptonlinedecisionmaking}, CAMEL~\citep{li2023camelcommunicativeagentsmind}, and AutoGen~\citep{dibia2024autogenstudionocodedeveloper} demonstrate that LLM agents can simulate diverse roles and complete complex tasks through dialogue-based coordination. A growing number of studies explore the use of MAS for translation-related tasks. Notably, MAS has emerged as a promising solution for document-level translation~\citep{wang2025deltaonlinedocumentleveltranslation}, a long-standing challenge in MT.

\paragraph{Formatted Text Translation.}
Formatted text translation concerns documents that contain structural or semantic markup, such as LaTeX and XML, where natural language content is interleaved with commands, tags, and tokens encoding layout or semantic information~\cite{kleidermacher,khan2025xmljson}. Research in this field faces two primary challenges.
Firstly, there is a lack of robust, general-purpose systems specifically designed for translating formatted documents. While several proprietary tools (\textit{e.g.}, Youdao and Baidu) provide practical solutions, open-source systems such as MathTranslate\footnote{\url{https://github.com/SUSYUSTC/MathTranslate}} and GPT-Academic\footnote{\url{https://github.com/binary-husky/gpt\_academic}} offer limited performance and scalability. 
Secondly, evaluation methodologies for formatted text translation remain inadequate. Conventional metrics such as BLEU \cite{papineni2002bleu} and COMET \cite{rei2020comet} focus on linguistic quality but fail to assess structural correctness and compilability. 

\section{System Design}
\label{sec:system}
The main architecture of LaTeXTrans is a multi-agent coordination designed for translating LaTeX source files. It consists of three modules: the Parser, the Translation Module, and the Generation Module. The design and functionality of each component are described in detail below.

\subsection{Parser Module}
\label{sec:parser}
Structured LaTeX documents interleave natural language content with formatting commands and semantic markup, resulting in tightly coupled representations that are not well-suited for direct translation by LLMs. Naively feeding the entire document to an LLM leads to several issues: unnecessary processing of non-translatable components, increased computational cost, and a higher risk of introducing translation errors.
To address these challenges, we introduce the Parser module, which serves as the first stage of the LaTeXTrans pipeline. Its basic idea is to transform complex LaTeX documents into clean, structured translation units that are easier for LLMs to process. Specifically, we design a placeholder substitution strategy to temporarily replace LaTeX-specific commands and environments, and implement a filtering mechanism to remove components that do not require translation.

\paragraph{Placeholder Substitution Strategy.}
\begin{figure*}[!t]
  \includegraphics[width=\textwidth]{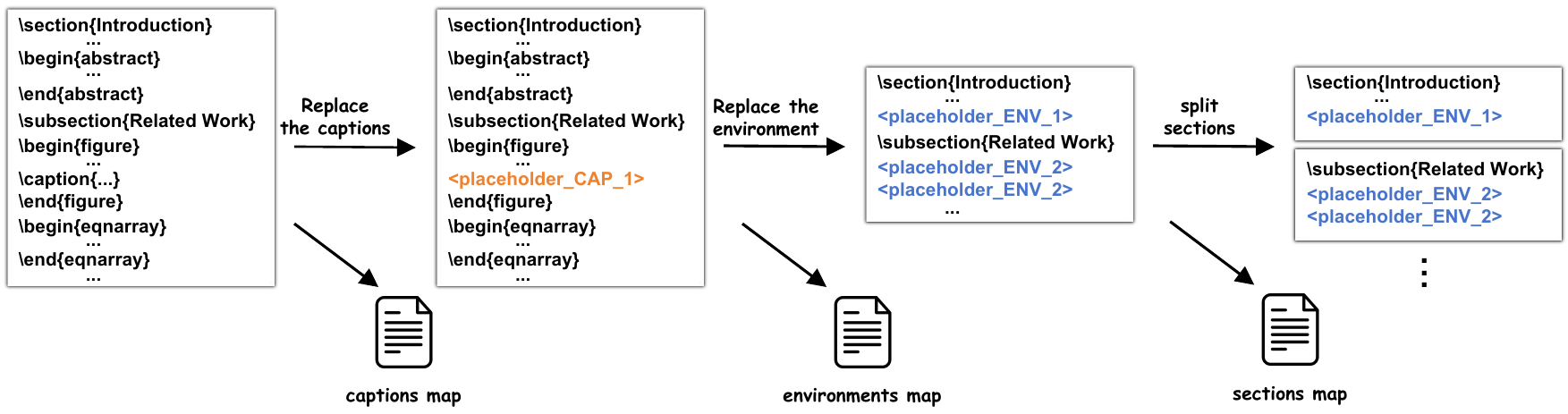}
  \vspace{-0.7cm}
  \caption{The pipeline of our placeholder substitution strategy. The mapping files are the mapping of placeholders and the replaced content, and they are also translation units of different granularities.}
  \vspace{-0.3cm}
  \label{fig:placeholder}
\end{figure*}

For a common LaTeX document, our placeholder substitution strategy is shown in Figure \ref{fig:placeholder}. We consider that the original mathematical formulas and charts are retained during translation. The first step is to replace the captions in the chart with placeholders. The second step is to replace the environment with placeholders, which will include the vast majority of mathematical formulas, charts, and other parts that do not need to be translated. Finally, we split the replaced text into sections (including subsections and subsubsections). For a LaTeX project composed of multiple tex files, we first merge the necessary tex files into the main file and then insert placeholders at the beginning and end of the merge for future restoration. The subsequent placeholder replacement rules and segmentation methods are the same as before. From the placeholder substitution strategy, we obtain translation units of two granularities: context (\textit{i.e.}, section and environment) and sentence (\textit{i.e.}, caption). 

\paragraph{Translation Unit Filter.}
While non-translatable components are replaced with placeholders, we notice that LaTeX allows users to define custom environments, making it infeasible to rely solely on exhaustive rule-based approaches to identify all such segments. To address this issue, we complement a predefined list of protected environments with a Filter agent powered by an LLM, which dynamically determines whether a given environment requires translation. Each extracted environment is annotated with a binary label: \texttt{True} or \texttt{False}. The translation module subsequently processes only those segments labeled as \texttt{True}.

\subsection{Translation Module}
The translation module comprises four agents: the Translator, Validator, Summarizer, and Terminology Extractor. After the Translator completes the translation of all designated units, the output is passed to the Validator, which generates an error report and returns it for revision if necessary. The Summarizer and Terminology Extractor assist the Translator by providing a summary of the preceding content and a domain-specific terminology dictionary, respectively, thereby enhancing contextual coherence and ensuring terminology consistency throughout the translation process.
\label{sec:transandval}

\paragraph{Translator-Validator Iteration.}
When utilizing large-context windows for document translation, large language models (LLMs) often prioritize capturing the overall meaning of the text, which can result in the omission or mistranslation of individual sentences~\cite {wang2025deltaonlinedocumentleveltranslation}. This issue is particularly pronounced in LaTeX document translation, where LLMs may neglect or incorrectly render LaTeX commands. For example, the command ``\texttt{\textbackslash textbf\{\}}'' may be omitted, or ``\texttt{\textbackslash left}'' may be incorrectly translated as ``\begin{CJK}{UTF8}{gbsn}\texttt{\textbackslash 左}\end{CJK}''.
Due to the structured and sensitive syntax of LaTeX, such errors are frequent and can lead to compilation failures. To address this issue, we introduce a Translator–Validator iterative framework, which performs multiple rounds of verification to progressively improve LaTeX command preservation for each translation unit. This iterative refinement significantly enhances the usability and reliability of the overall translation system. Specifically, as illustrated in Figure \ref{fig:LaTeXTrans}, after the Translator has completed the translation of all translation units, the Validator will verify the quality of the translation from three dimensions and eventually generate an error report. When conducting the next round of translation, the erroneous translation units, together with the error reports, will form the prompt for the Translator to guide them in generating the correct translation. 

\paragraph{Summarizer and Terminology Extractor.}
Inspired by \citet{wang2025deltaonlinedocumentleveltranslation}'s work, we design a Summarizer and Terminology Extractor to enhance the contextual coherence and terminology consistency of translation. Specifically, the Summarizer is responsible for constantly generating and updating the summary of the previous text during the translation process. When each translation unit is completed, the Summarizer will combine the previous summary with the original text of the current translation unit to generate a new summary. The Terminology Extractor is responsible for maintaining a terminology dictionary and adding it to the prompt of the Translator to provide a reference for terminology translation for the Translator. When the Translator finishes the translation of a translation unit, the Terminology Extractor extracts term pairs from the original text and the translation and updates the term dictionary in real-time.

\subsection{Generation Module}
The generation module automatically selects the appropriate compilation engine (\textit{e.g.}, pdf\LaTeX{} or Xe\LaTeX{}) based on the document characteristics, and subsequently compiles the structured and translated LaTeX source files into PDF outputs.

\begin{figure*}[t]
  \includegraphics[width=\textwidth]{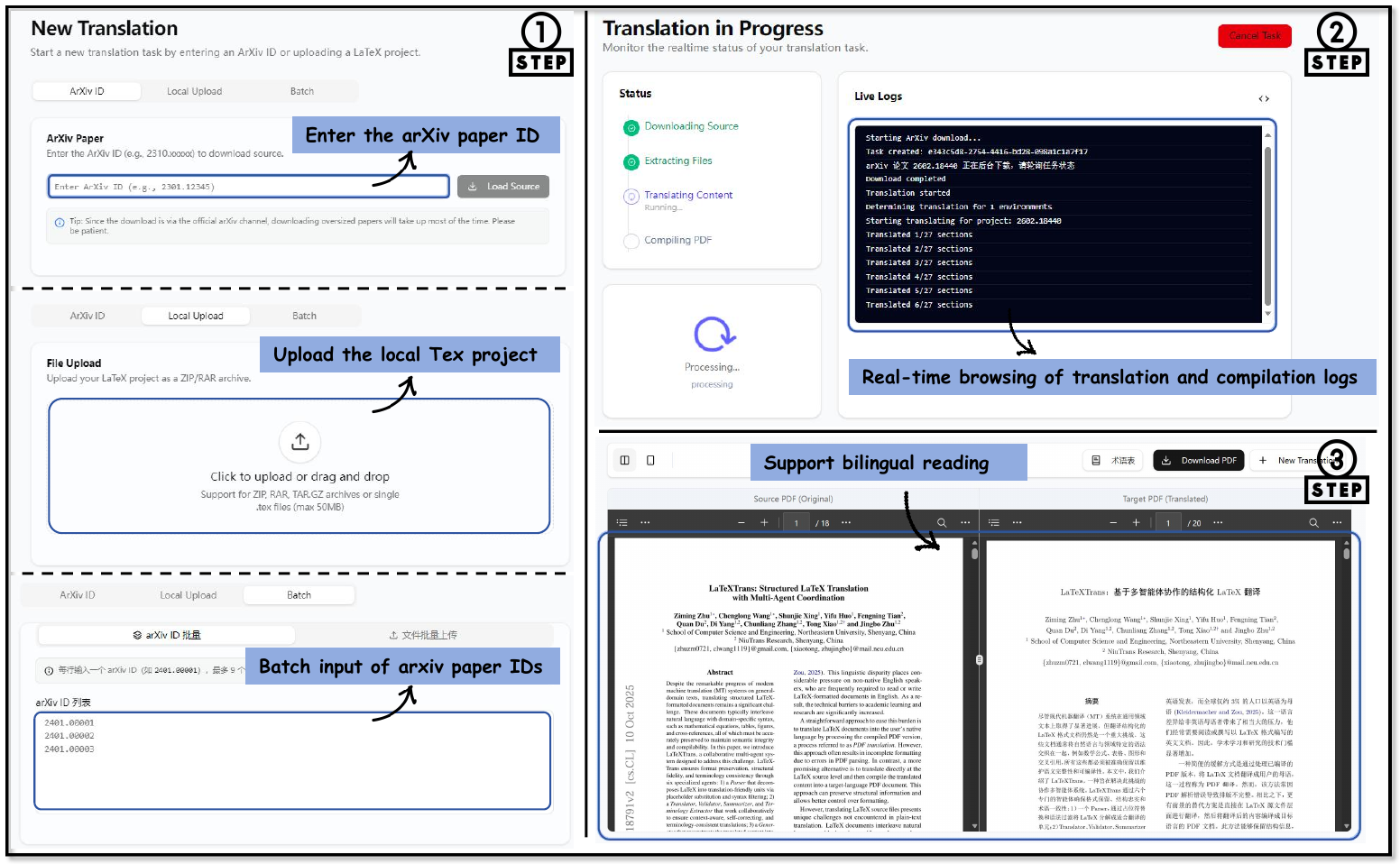}
  \vspace{-0.7cm}
  \caption{
    Illustration of the LaTeXTrans online platform interface and workflow.
  }
  \vspace{-0.2cm}
  \label{fig:web}
\end{figure*}

\section{System Deployment and Usage}
\subsection{Command-Line Tool}
LaTeXTrans supports local deployment through a command-line interface. After configuring the API credentials, we can run fully automated end-to-end translation. By providing an arXiv paper ID, the system retrieves the associated LaTeX project and produces a translated PDF in a single command:
\begin{center}
\ttfamily\small LaTeXTrans --arxiv xxxx.xxxxx
\end{center}
Beyond API setup, LaTeXTrans supports flexible customization, including target language selection and the integration of user-defined terminology dictionaries. Upon completion, the system generates both the translated PDF and the complete translated LaTeX source project. Additionally, detailed translation and compilation logs are provided to facilitate debugging and ensure transparency.
\subsection{Online Platform}
To further enhance accessibility, we provide a web-based platform built upon LaTeXTrans for LaTeX document translation. The platform supports three input modes: (1) automatic retrieval via arXiv paper IDs, (2) batch processing of multiple IDs, and (3) uploading local LaTeX projects. During execution, users can monitor real-time translation and compilation logs. Additionally, the platform offers bilingual PDF comparison to facilitate side-by-side reading. The usage workflow of the online platform is illustrated in Figure \ref{fig:web}.

\section{Experiment}
\label{sec:experiment}

\subsection{Settings}
\paragraph{Datasets.}
As no publicly available benchmark currently exists for LaTeX source-level document translation, we construct a test set comprising the LaTeX source files of 100 English arXiv papers\footnote{Although the test set contains 100 papers, this scale is sufficient for LaTeX-based evaluation, as each paper typically spans multiple pages and contains rich structural elements, resulting in substantial document-level complexity.}. The dataset includes 50 papers from computer science, 30 from physics, and 20 from mathematics. To further stress-test the system, we deliberately include 10 long survey papers characterized by complex layouts and extended document structures. This design ensures sufficient scale, domain diversity, and structural complexity, enabling a comprehensive and realistic evaluation of our approach.

\paragraph{Baselines.}
To comprehensively evaluate both translation quality and format preservation, we categorize the baselines into two groups: section-level direct translation and formatted-text translation systems. For the former, we selected NiuTrans and Google Translate as traditional MT systems, and Qwen-3-14B \cite{yang2025qwen3}, DeepSeek-V3 \cite{liu2024deepseek}, and GPT-4o \cite{hurst2024gpt} as LLM-based translation systems. For the latter, we selected gpt-academic as the baseline. When evaluating section-level direct translation systems, we segment LaTeX documents into section-level units and translate them directly without structural parsing or iterative validation. 


\begin{table}[!t]
  \vspace{1.5mm}
  \centering
  \resizebox{\linewidth}{!}{
  \input{tables/format_res}}
  \vspace{-2mm}
    \caption{
    The performance of LaTeXTrans in three translation tasks for papers from different fields.
    }
    \vspace{-4mm}
\label{tab:SR result}
\end{table}

\subsection{Main Results}
\paragraph{Structural Preservation.}
To evaluate the structural performance of LaTeXTrans, we report the Compilation Success Rate (CSR) and the Formatting Error Count (FEC). CSR measures the percentage of translated LaTeX projects that compile successfully without manual intervention, while FEC represents the average number of formatting errors per document. Detailed error categories and representative examples are provided in the Appendix~\ref{app:Errors}. We evaluate LaTeXTrans on three translation tasks: English-Chinese (En–Zh), English–Japanese (En–Ja), and English–Korean (En–Ko). As shown in Table~\ref{tab:SR result}, we find that LaTeXTrans can achieve a CSR of 97\% in the En–Zh setting and consistently high CSR scores in the En–Ja and En–Ko tasks, demonstrating strong structural reliability across diverse LaTeX projects. Notably, CSR reaches 100\% in the physics and mathematics domains. Beyond compilation success, LaTeXTrans can maintain consistently low FEC values across all tasks. Note that the slightly higher FEC observed in mathematics is primarily due to the dense and complex mathematical environments, which increase sensitivity to formatting inconsistencies during reconstruction. 


\begin{table}[t]
  \centering
  \resizebox{\linewidth}{!}{
  \input{tables/quality_res}}
\vspace{-0.3cm}
\caption{
COMETkiwi and LLM-score comparisons across different systems. 
Bold indicates the best result in each group.
}
\label{tab:TQ result}
\end{table}

\begin{table}[!t]
  \centering
  \resizebox{\linewidth}{!}{\input{tables/humen_eval}}
\vspace{-0.3cm}
\caption{
Human evaluation of format preservation (En–Zh). A: perfect preservation; B: minor format errors; C: severe format corruption.
}
\vspace{-4mm}
\label{tab:HM-results}
\end{table}

\paragraph{Translation Quality.}
We evaluate LaTeXTrans using COMETkiwi~\cite{rei2022cometkiwiistunbabel2022submission} and LLM-score. Detailed evaluation settings are provided in Appendix~\ref{app:evaluation}. The results are listed in Table~\ref{tab:TQ result}. First, compared with traditional MT systems, LaTeXTrans demonstrates clear performance advantages, highlighting the effectiveness of our LaTeX-specific optimization strategies. When compared to LLM-based translation approaches, our system achieves comparable performance while preserving structural fidelity, and even surpasses them on certain metrics. We attribute these gains to our fine-grained parsing strategy and placeholder-based isolation of non-textual elements. 

\paragraph{Human Evaluation.}
To complement automatic structural metrics, we conduct a human evaluation on the En–Zh translation task to assess format preservation quality. Translated documents are categorized into three levels based on formatting quality (see Appendix~\ref{app:detailed-settings} for details): (A) perfect preservation of the original format, (B) minor formatting errors that do not affect readability, and (C) severe errors that significantly impair reading. As shown in Table~\ref{tab:HM-results}, LaTeXTrans achieves a substantially higher proportion of level-A outputs compared to GPT-Academic, demonstrating superior format preservation. These findings indicate that LaTeXTrans not only improves automatic structural metrics but also delivers visibly higher formatting quality from a human evaluation.

\subsection{Ablation Study}
Table~\ref{tab:ablation} presents an ablation study on the En–Zh task. Introducing the Parser module significantly improves COMETkiwi, indicating that the placeholder substitution strategy enhances translation quality. Adding the Validator module further boosts overall performance, although a slight drop in LLM-score is observed with DeepSeek-V3. We hypothesize that this is due to the Validator enforcing strict tag retention through iterative checks, which may restrict the Translator and slightly impact fluency. Finally, incorporating the Summarizer and Terminology Extractor improves the LLM-score, reflecting better cross-paragraph coherence. 

\begin{table}[t]
  \setlength{\tabcolsep}{1.5pt}
  \small
  \centering
  \resizebox{\columnwidth}{!}{
\input{tables/ablation}
  }
  \vspace{-2mm}
  \caption{
  Ablation study of LaTeXTrans.
  ``SA.'' denotes the LLM-based translation baseline, ``P.'' stands for the Parser, ``V.'' for the Validator, ``S.'' for summarizer, and ``TE.'' for the Terminology Extractor. The ``SA. + P. + V. + S. + TE.'' corresponds to our LaTeXTrans.
  }
  \vspace{-0.4cm}
  \label{tab:ablation}
\end{table}


\section{Conclusion}
In this paper, we propose LaTeXTrans, a multi-agent system for translating structured LaTeX documents. LaTeXTrans consists of three collaborative modules, each responsible for a specific stage of the translation pipeline. Experimental results demonstrate that LaTeXTrans can outperform baseline systems and offer a reliable solution for LaTeX document translation.

\section*{Limitations}
Any instruction-following LLM can be integrated into our LaTeXTrans system. However, due to the large number of available models, it is impractical to evaluate each one individually. Therefore, we select a representative subset of commonly used LLMs for our experiments. We believe this selection sufficiently demonstrates the practicality and effectiveness of LaTeXTrans for LaTeX document translation. Additionally, although commercial systems such as Baidu and Youdao offer LaTeX translation services, they are not open-source. As a result, we are unable to compute metrics like COMETkiwi and LLM-score for these systems. Therefore, we do not include a comprehensive comparison with them in our main experiments.

\bibliography{custom}

\appendix
\clearpage


\section{Additional Detailed Settings of the Experiment}
\label{app:detailed-settings}

\paragraph{Baselines.}
Since the dataset consisted entirely of structured LaTeX documents which exceeded the handling capabilities of single-model systems, we adopted a preprocessing step in the baseline approach. Specifically, the structured LaTeX documents were segmented into section-level translation units to make them manageable for translation.

\paragraph{Hyperparameter Setting.}
In the experiments, we evaluated both open-source and closed-source models separately. For the closed-source models, we accessed them via a third-party API. In the baseline approach, we set the maximum number of new tokens to 16,384 and the temperature to 0.7, while keeping all other hyperparameters at their default values. For our system, the temperature in the Filter was set to 0 with a maximum of 50 new tokens, while all other agents were configured with a maximum of 8,192 new tokens; the remaining hyperparameters were kept at their defaults.

\paragraph{Evaluation.}
\label{app:evaluation}
We use GPT-4o to calculate LLM-scores. When computing COMETkiwi and LLM-scores, we used \texttt{pylatexenc}\footnote{\url{https://github.com/phfaist/pylatexenc}} to convert each LaTeX translation unit into plain text. Although LaTeXTrans parses structured LaTeX documents into fine-grained translation units, we followed the baseline's evaluation protocol by using section-level translation units for computing both COMETkiwi and LLM-scores. Furthermore, to assess contextual consistency in the LLM-score evaluation, we concatenated section-level translation units into paired paragraphs and then scored them using GPT-4o. 

\begin{figure}[!t]
\centering
  \includegraphics[width=\columnwidth]{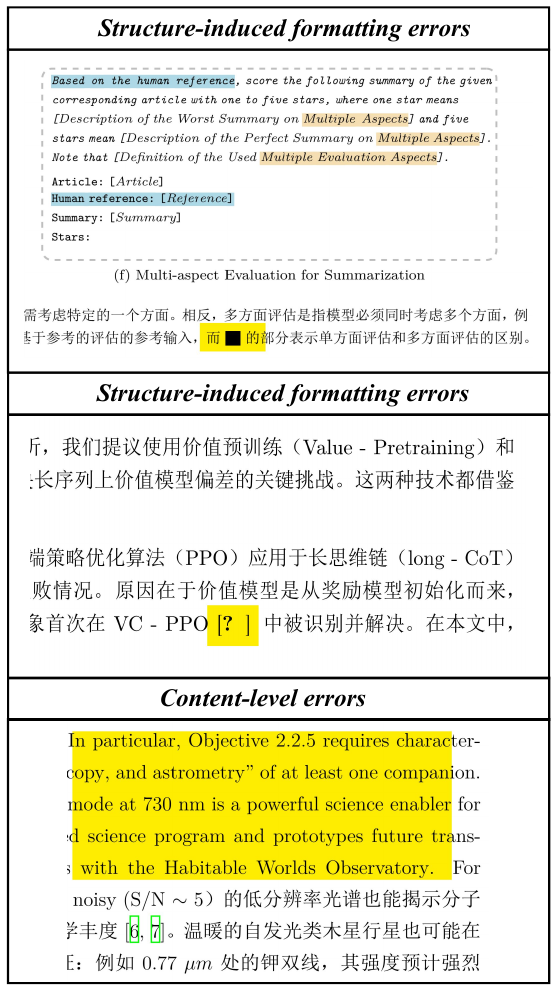}
  \vspace{-0.7cm}
  \caption{
    Examples of formatting errors.
  }
  \vspace{-0.2cm}
  \label{fig:error1}
\end{figure}

\section{Formatting Error Categories}
\label{app:Errors}
We categorize formatting-related errors into two major types:
\vspace{-2mm}
\begin{itemize}
    \item \textit{Structure-induced formatting} errors, arising from corrupted or improperly restored \LaTeX{} commands. These may result in missing formatting effects (e.g., bold text not rendered), incorrect cross-references, or garbled content in the compiled document.
    \vspace{-2mm}
    \item \textit{Content-level errors}, involving incorrect or unintended textual outputs, such as untranslated segments or irrelevant words.
\end{itemize}
\vspace{-2mm}
Figure~\ref{fig:error1} illustrates representative examples of formatting errors.

To ensure consistent evaluation, we adopt the following counting rules. For structure-induced errors, if a single corrupted \LaTeX{} command causes multiple downstream anomalies (e.g., numerous missing cross-references triggered by one faulty reference definition), the error is counted only once per document. In contrast, content-level errors are counted independently at the instance level. This design prevents inflated error counts caused by cascading structural failures while preserving fine-grained sensitivity to semantic and textual inaccuracies.

\section{Case Study}
\label{app:Case Studies}
\begin{figure*}[!t]
    \centering %
    \begin{subfigure}[t]{0.48\textwidth}
        \centering
        \includegraphics[width=\linewidth]{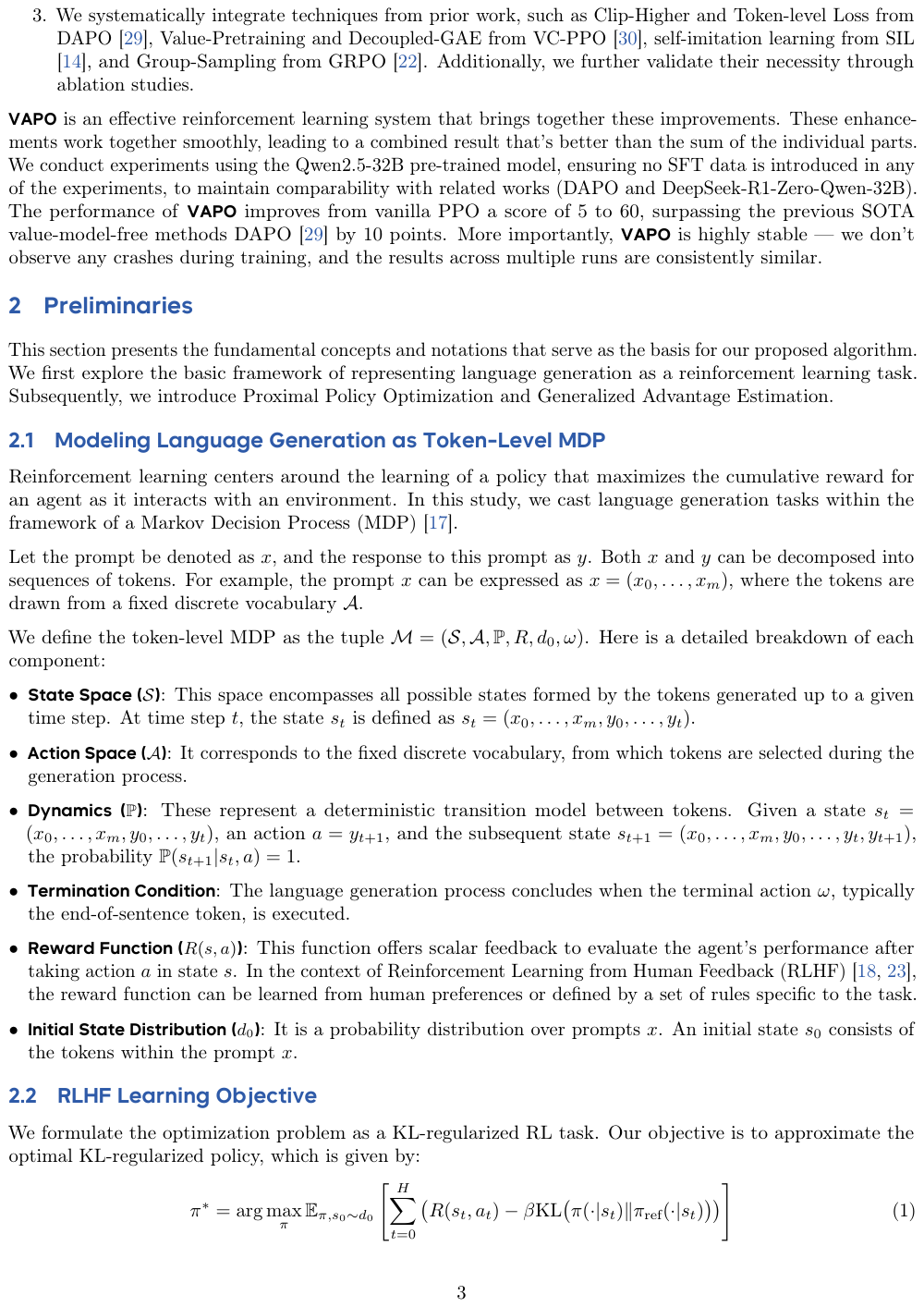}
        \subcaption{Part of the English PDF of case 1.} %
        \label{fig:source1}
    \end{subfigure}
    \hfill %
    \begin{subfigure}[t]{0.48\textwidth}
        \centering
        \includegraphics[width=\linewidth]{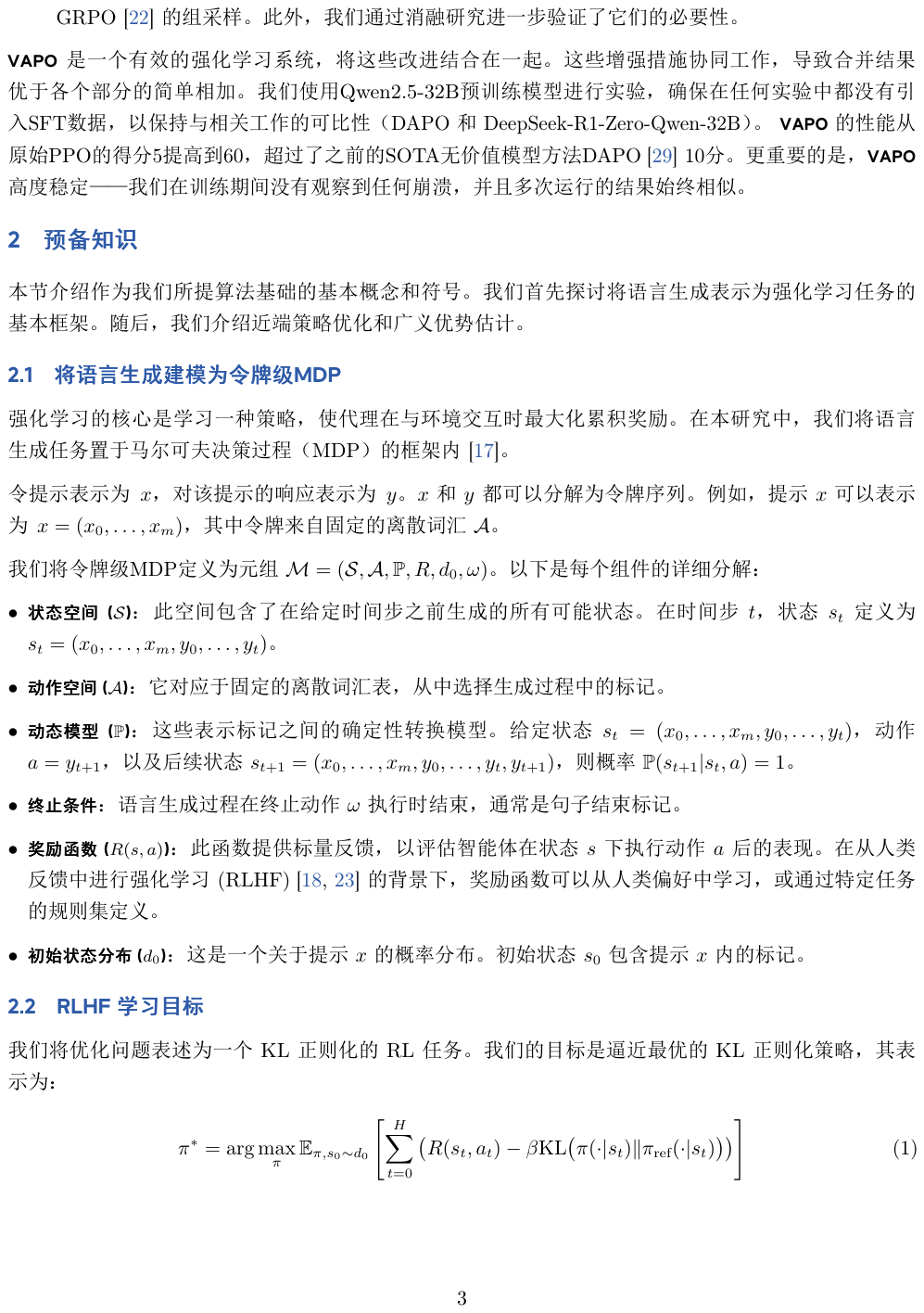}
        \subcaption{Part of the Chinese PDF of case 1.} %
        \label{fig:trans1}
    \end{subfigure}

    \begin{subfigure}[t]{0.48\textwidth}
        \centering
        \includegraphics[width=\linewidth]{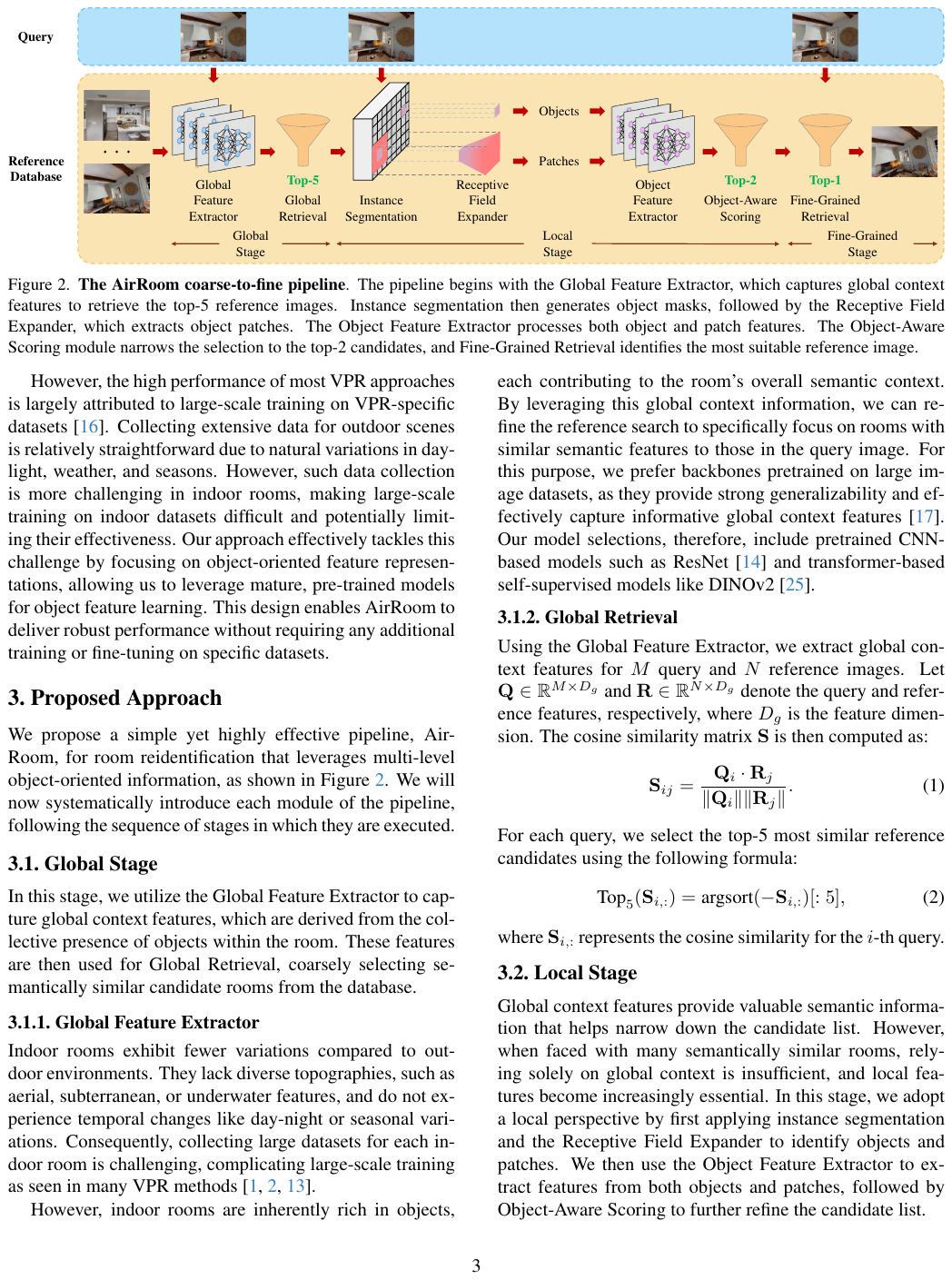}
        \subcaption{Part of the English PDF of case 2.} %
    \end{subfigure}
    \hfill %
    \begin{subfigure}[t]{0.48\textwidth}
        \centering
        \includegraphics[width=\linewidth]{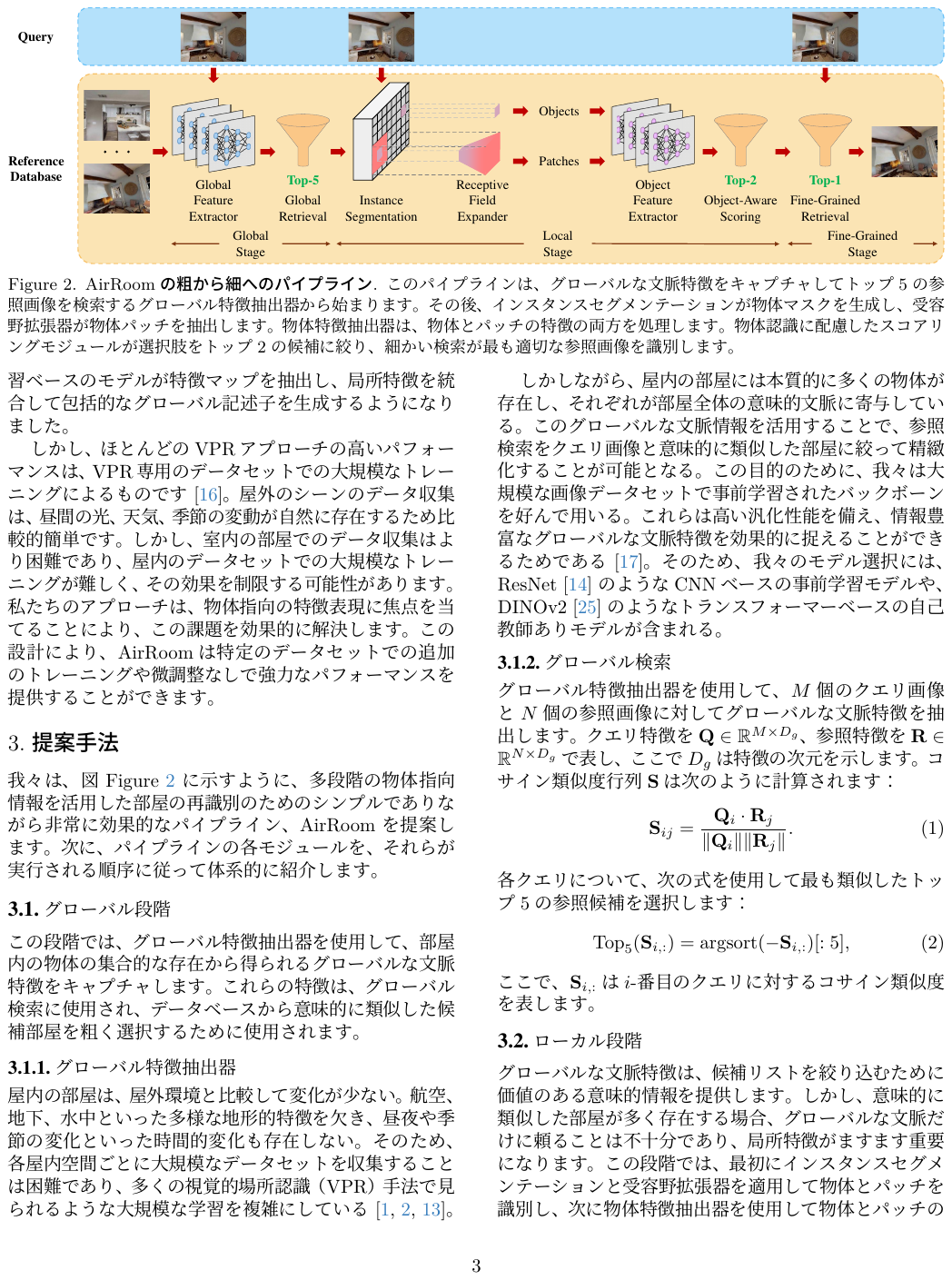}
        \subcaption{Part of the Japaese PDF of case 2.} %
    \end{subfigure}

    \caption{Two cases demonstrate the performance of LaTeXTrans on the En-Zh and En-Ja task} %
    \label{fig:case1}
\end{figure*}

We present two case studies to visually demonstrate the translation performance of LaTeXTrans on the En–Zh and En–Ja tasks, as shown in Figure~\ref{fig:case1}. Each case corresponds to the translation of a real-world LaTeX source document. For illustration, we highlight two relatively complex segments in each example.

\clearpage
\par

\end{document}

%% file: tables/format_res.tex
  \begin{tabular}{lcccccc}
  \toprule[1.1pt]
  \multirow{2}{*}{\textbf{Subject}} & 
  \multicolumn{2}{c}{\textbf{En-Zh}}  & 
  \multicolumn{2}{c}{\textbf{En-Ja}} &
  \multicolumn{2}{c}{\textbf{En-Ko}}
  \\
  \cmidrule(l){2-3} \cmidrule(l){4-5} \cmidrule(l){6-7}
  &   \bf CSR ($\uparrow$) & \bf FEC ($\downarrow$) & \bf CSR ($\uparrow$)  & \bf FEC ($\downarrow$)  & \bf CSR ($\uparrow$) & \bf FEC ($\downarrow$)  \\
    \midrule
    Physics         & 100\%     &  0.37      & 93\%  & 0.43     & 90\%    &  0.47      \\
    Mathematics     & 100\%     & 1.15      & 95\%   & 1.30  &  100\%   &    1.35   \\
    Computer Science    & 94\%    &  0.18     & 76\%   & 0.20   & 86\%   & 0.25     \\
  \midrule
    Total    & 97\%    &  0.43     & 85\%   & 0.49  & 90\%   & 0.54     \\
  \bottomrule[1.1pt]
\end{tabular}

%% file: tables/quality_res.tex
  \begin{tabular}{lcccc}
  \toprule[1.1pt]
  \multirow{2}{*}{\textbf{System}} & 
  \multicolumn{2}{c}{\textbf{En-Zh}}  & 
  \multicolumn{2}{c}{\textbf{En-Ja}} 
  \\
  \cmidrule(l){2-3} \cmidrule(l){4-5} 
  &   \bf Cometkiwi   &  \bf LLM-score   &   \bf Cometkiwi  &  \bf LLM-score   \\
    \midrule
    NiuTrans               &64.69        &7.93   &65.49                &8.19      \\
    Google Translate                           &46.23       &5.93   &56.21       &7.01     \\
    \hdashline
    Qwen-3-14b      &68.18 & 8.76 & 72.84 & 8.66 \\
    LaTeXTrans \textsubscript{Qwen-3-14b}            & 71.37     & 8.97      & 74.68   & 8.51        \\
    \hdashline
    DeepSeek-V3       &67.26  & \textbf{9.02}  & 72.17 & \textbf{9.00} \\
    LaTeXTrans \textsubscript{DeepSeek-V3}       &73.48      & 9.01      & \textbf{75.39}  &  8.89           \\
    \hdashline
    GPT-4o          &67.22  & 8.58 & 71.16 &8.91 \\
    LaTeXTrans \textsubscript{GPT-4o}             & \textbf{73.59}     & 8.92     & 74.47   &  8.93       \\
  \bottomrule[1.1pt]
\end{tabular}

%% file: tables/humen_eval.tex
\begin{tabular}{lc*{8}{>{\centering\arraybackslash}p{0.05\textwidth}}}
  \toprule[1.1pt]
  \multirow{2}{*}{\textbf{System}} & 
  \multirow{2}{*}{\textbf{Total (70)}} & 
  \multicolumn{3}{c}{\textbf{Mathematics (20)}} &
  \multicolumn{3}{c}{\textbf{Computer Science (50) }}
  \\
  \cmidrule(l){3-5}   \cmidrule(l){6-8}  
  & &   A &  B &  C  &  A & B & C     \\
    \midrule
    LaTeXTrans   & 67      &14 &3 &3  &45 &1 &1                \\
    gpt-academic  & 69             &9 &6 &5  &40 & 7&   2         \\
  \bottomrule[1.1pt]
\end{tabular}

%% file: tables/ablation.tex
  \begin{tabular}{lcccc}
  \toprule[1.1pt]
  \multirow{2}{*}{\textbf{Setting}} & \multicolumn{2}{c}{\textbf{GPT-4o}}  & \multicolumn{2}{c}{\textbf{DeepSeek-V3}} \\
  \cmidrule(r){2-3} \cmidrule(r){4-5}
  & \bf Cometkiwi & \bf LLM-score &  \bf Cometkiwi & \bf LLM-score   \\
  \midrule
  SA. (Baseline)                & 67.22 & 8.58   & 67.26 & 9.02   \\  \hdashline
  SA. + P.           & 74.47 & 8.89   & 74.39 & 9.03   \\
  SA. + P. + V.      & \textbf{74.57} & 8.91     & \textbf{74.42} & 8.94   \\
  SA. + P. + V. + S. & 74.06  & \textbf{8.95}    & 74.02 & \textbf{9.05}   \\
  SA. + P. + V. + S. + TE.  & 73.59 & 8.93   & 73.48 & 9.01   \\
  \bottomrule[1.1pt]
  \end{tabular}